\documentclass[10pt,twocolumn,letterpaper]{article}

\usepackage[pagenumbers]{cvpr} % To force page numbers, \eg, for an arXiv version

\usepackage{graphicx}
\usepackage{amsmath}
\usepackage{amssymb}
\usepackage{booktabs}

\usepackage{multirow}
\usepackage{pifont}% http://ctan.org/pkg/pifont
\usepackage[dvipsnames]{xcolor}
\usepackage{bbm}

\usepackage{times}
\usepackage{epsfig}
\usepackage{graphicx}
\usepackage{float}
\usepackage{wrapfig}
\usepackage{amsmath,amssymb,amsthm}
\usepackage{algorithm,algorithmicx,algpseudocode}
\usepackage{bm,xspace}
\usepackage{comment}
\usepackage{verbatim}
\usepackage{multirow}
\usepackage{balance}
\usepackage{url}
\usepackage{booktabs}
\usepackage{etoolbox}
\usepackage{siunitx}
\usepackage{calc}
\usepackage{pifont,hologo}
\usepackage{nicefrac}
\usepackage[normalem]{ulem}

\usepackage[super]{nth}
\usepackage{nicefrac}
\robustify\bfseries
\robustify\uline

\DeclareMathSymbol{@}{\mathord}{letters}{"3B}

\newcommand\mypara[1]{\vspace{1mm}\noindent\textbf{#1}}

\def\latex/{\LaTeX}
\def\bibtex/{\hologo{BibTeX}}

\definecolor{mygreen}{RGB}{0, 102, 0}
\definecolor{myred}{RGB}{153, 0, 0}

\usepackage[pagebackref,breaklinks,colorlinks]{hyperref}

\usepackage[capitalize]{cleveref}
\usepackage{dashrule}
\crefname{section}{Sec.}{Secs.}
\Crefname{section}{Section}{Sections}
\Crefname{table}{Table}{Tables}
\crefname{table}{Tab.}{Tabs.}

\begin{document}

%%%%%%%%% TITLE - PLEASE UPDATE
\title{MiDaS v3.1 -- A Model Zoo for Robust Monocular Relative Depth Estimation}

\author{Reiner Birkl, Diana Wofk, Matthias M{\"u}ller\\
\vspace{-5pt}\\
Intel Labs\\
% Institution1 address\\
% {\tt\small firstauthor@i1.org}
% For a paper whose authors are all at the same institution,
% omit the following lines up until the closing ``}''.
% Additional authors and addresses can be added with ``\and'',
% just like the second author.
% To save space, use either the email address or home page, not both
% \and
% Second Author\\
% Institution2\\
% First line of institution2 address\\
% {\tt\small secondauthor@i2.org}
}

\maketitle

%%%%%%%%% ABSTRACT
\begin{abstract}

We release MiDaS v3.1\footnote{\href{https://github.com/isl-org/MiDaS}{github.com/isl-org/MiDaS}} for monocular depth estimation, offering a variety of new models based on different encoder backbones. This release is motivated by the success of transformers in computer vision, with a large variety of pretrained vision transformers now available. We explore how using the most promising vision transformers as image encoders impacts depth estimation quality and runtime of the MiDaS architecture. Our investigation also includes recent convolutional approaches that achieve comparable quality to vision transformers in image classification tasks. While the previous release MiDaS v3.0 solely leverages the vanilla vision transformer ViT, MiDaS v3.1 offers additional models based on BEiT, Swin, SwinV2, Next-ViT and LeViT. These models offer different performance-runtime trade-offs. The best model improves the depth estimation quality by $28\%$ while efficient models enable downstream tasks requiring high frame rates. We also describe the general process for integrating new backbones.
\end{abstract}

%%%%%%%%% BODY TEXT
\section{Introduction}
\label{sec:intro}
Monocular depth estimation refers to the task of regressing dense depth solely from a single input image or camera view. Solving this problem has numerous applications in downstream tasks like generative AI~\cite{rombach2021highresolution, zhang2023adding, hollein2023text2room}, 3D reconstruction~\cite{mildenhall2021nerf, hu2023consistentnerf, chen2023improving} and autonomous driving~\cite{liu2022vision, fonder2021m4depth}. However, it is particularly challenging to deduce depth information at individual pixels given just a single image, as monocular depth estimation is an under-constrained problem. Significant recent progress in depth estimation can be attributed to learning-based methods. In particular, dataset mixing and scale-and-shift-invariant loss construction has enabled robust and generalizable monocular depth estimation with MiDaS~\cite{ranftl2020towards}. Since the initial development of that work, there have been several releases of MiDaS offering new models with more powerful backbones\cite{Ranftl_2021_ICCV_DPT} and lightweight variants for mobile applications. 

Many deep learning models for depth estimation adopt encoder-decoder architectures. In addition to convolutional encoders used in the past, a new category of encoder options has emerged with transformers for computer vision. Originally developed for natural language processing~\cite{vaswani2017attention} and nowadays the foundation of large language models like ChatGPT~\cite{liu2023summary}, transformers have led to a wide variety of new vision encoders since the first vision transformer ViT~\cite{vit}. Many of these new encoders have surpassed the performance of previous convolutional encoders. Inspired by this, we have identified the most promising transformer-based encoders for depth estimation and incorporated them into MiDaS. Since there have also been attempts to make convolutional encoders competitive ~\cite{convnext, efficientnetl2, yu2022metaformer}, we also include these for a comprehensive investigation.

The latest release MiDaS v3.1, which is the focus of this paper, offers a large collection of new depth estimation models with various state-of-the-art backbones. The goal of this paper is to describe the integration of these backbones into the MiDaS architecture, to provide a thorough comparison and analysis of the different v3.1 models available, and to provide guidance on how MiDaS can be used with future backbones.

\section{Related Work}

Monocular depth estimation is inherently an ill-posed problem facing challenges like metric scale ambiguity. Learning-based approaches that aim to directly regress metric depth~\cite{bhat2021adabins, bhat2022localbins, jun2022depth, li2022binsformer, yuan2022new} have sought to use supervised training on homogeneous datasets with representative environments (\eg, focusing on indoor or outdoor scenes) to encourage the supervised network to learn an appropriate metric scale. However, this results in overfitting to narrow depth ranges and degrades generalizability across environments. Alternatively, relative depth estimation (RDE) approaches~\cite{lee2019monocular, ranftl2020towards, Ranftl_2021_ICCV_DPT} aim to regress pixel-wise depth predictions that are accurate relative to each other but carry no metric meaning. The scale factor and potentially a shift factor remain unknown. By factoring out metric scale, these RDE approaches are able to be supervised through disparity labels, which allows training on combinations of heterogeneous datasets with varying metric depth scales and camera parameters. This enables improved model generalizability across environments.

The MiDaS family of models originates from a key work in the relative depth estimation space that demonstrated the utility of mixing datasets to achieve superior zero-shot cross-dataset performance~\cite{ranftl2020towards}. Depth prediction is performed in disparity space (\ie, inverse depth up to scale and shift), and training leverages scale-and-shift-invariant losses to handle ambiguities in ground truth labels. Existing depth estimation datasets are mixed together and complemented with frames and disparity labels from 3D movies, thus forming a large meta-dataset. As MiDaS releases have progressed through several versions, more datasets have been incorporated over time. Datasets are discussed as part of the training overview in~\cref{sec:training_setup}.

The network structure of MiDaS follows a conventional encoder-decoder structure, where the encoder is based on an image-classification network. The original MiDaS v1.0 and v2.0 models use the ResNet-based~\cite{He_2016_CVPR} multi-scale architecture from Xian~\etal~\cite{xian2018monocular}. A mobile-friendly variant using an EfficientNet-Lite~\cite{Tan2019EfficientNetRM} backbone is released as part of MiDaS v2.1. Transformer-based backbones are explored in MiDaS v3.0\cite{Ranftl_2021_ICCV_DPT}, where variants of ViT~\cite{vit} are integrated into the MiDaS architecture to develop {Dense Prediction Transformers}~\cite{Ranftl_2021_ICCV_DPT}. This report follows up on these efforts by demonstrating how newer backbones, both convolutional and transformer-based, can be integrated into MiDaS, as well as how depth estimation performance benefits from these novel encoder backbones. Our new models are released as MiDaS v3.1.

\section{Methodology}

In this section, we first provide a detailed overview of convolutional and transformer-based backbones that we explore when developing models for MiDaS v3.1. We then explain how these encoder backbones are integrated into the MiDaS architecture. Lastly, we describe the training setup and discuss a general strategy for adding new backbones for future extensions.

\subsection{Overview of Encoder Backbones}
A key guideline for the exploration of new backbones is that the depth estimation quality and compute requirements of alternative encoders in the MiDaS~\cite{ranftl2020towards} architecture should roughly correlate to their behavior in the original task, which is typically image classification. High quality and low compute requirements are generally mutually exclusive. To cover both tradeoffs for downstream tasks, we have implemented and validated different types of encoders which either provide the highest depth estimation quality or need the least resources.

\subsubsection{Published Models}
For the release of MiDaS v3.1, we have selected the five encoder types which seem most promising for downstream tasks, either due to their high depth estimation quality or low compute requirements for real time applications. This selection criterion also holds for the different sizes which are usually available for encoder types, like small and large. Our overview therefore splits into three parts: models with new backbones which are part of the MiDaS v3.1 release, models with backbones which were explored but not released and for completeness also the models of earlier MiDaS versions, because some of them are included as legacy models in MiDaS v3.1.

We begin with the new backbones released in MiDaS v3.1, which are all transformer backbones. The highest depth estimation quality is achieved with the BEiT~\cite{beit1} transformer, where we offer the BEiT\textsubscript{512}-L, BEiT\textsubscript{384}-L and BEiT\textsubscript{384}-B variants. The numbers denote the quadratic training resolutions 512x512 and 384x384, while the letters L and B stand for large and base. The BEiT transformer architecture also offers two newer versions, but we did not explore BEiT v2~\cite{beit2} and BEiT-3~\cite{beit3}. For BEiT v2~\cite{beit2} no pretrained checkpoint with a resolution of 384x384 or higher was available, but only checkpoints at 224x224. BEiT-3~\cite{beit3} was released after we completed the study.

The encoder type yielding the second highest depth estimation quality is the Swin transformer, where we offer models with both Swin~\cite{swin} and SwinV2~\cite{swin2} backbones. The available variants with high depth estimation quality are Swin-L, SwinV2-L and SwinV2-B, which are all at the resolution 384x384. For downstream tasks with low compute resources, we also offer a model based on SwinV2-T, with the resolution 256x256 and T denoting tiny. A characteristic of the MiDaS v3.1 models based on the Swin and SwinV2 transformer backbones as provided by the PyTorch Image Models repository~\cite{rw2019timm} is that only quadratic inference resolutions can be used. This is different to other newly released models where the inference resolution may differ from the training resolution.

The last two encoder types released in MiDaS v3.1 are Next-ViT~\cite{nextvit} as well as LeViT~\cite{levit} for low compute downstream tasks. For Next-ViT, we offer a model based on the Next-ViT-L ImageNet-1K-6M encoder at resolution 384x384. For LeViT, there is the variant LeViT-384 at resolution 224x224, which can be used at only quadratic inference resolutions like the Swin transformers. Note that according to the naming convention of the LeViT paper~\cite{levit} the number 384 in the transformer model name LeViT-384 does not stand for the training resolution but the number of channels in the first stage of the LeViT architecture. As we follow the convention that MiDaS models use the training resolution in the model names, the MiDaS model based on the transformer backbone LeViT-384 is called LeViT\textsubscript{224}.

\subsubsection{Unpublished Models}
Next, we give an overview of the backbones explored when developing MiDaS v3.1 that were ultimately rejected due to the resulting depth estimation models being less competitive. This overview includes both transformer and convolutional backbones. For the transformer backbones, we first come back to Next-ViT~\cite{nextvit}, where we have also tested Next-ViT-L ImageNet-1K. Our exploration also contains a variant of the vanilla vision transformer, which is ViT-L Hybrid. The next type of transformer is DeiT3~\cite{deit3}, where we have explored vanilla DeiT3-L as well as DeiT3-L pretrained on ImageNet-22k and fine-tuned on ImageNet-1K. All these four transformer backbones are at the resolution 384x384. Finally, there is MobileViTv2~\cite{mehta2022mobilevitv2} for less powerful hardware, where we have implemented the smallest variant MobileViTv2-0.5 at the resolution 256x256 and the largest one, MobileViTv2-2.0 at 384x384. The latter is pretrained on ImageNet-22K and fine-tuned on ImageNet-1K. The numbers 0.5 and 2.0 in the transformer names refer to the width multiplier used in the MobileViTv2 architecture.

We proceed with exploring convolutional backbones, where we consider ConvNeXt~\cite{convnext} and EfficientNet~\cite{efficientnetl2}. For ConvNeXt, we have implemented two variants pretrained on ImageNet-22K and fine-tuned on ImageNet-1K, which are ConvNeXt-L and ConvNeXt-XL. For EfficientNet~\cite{tan2019efficientnet}, we did not consider any of the base variants EfficientNet-B0 to EfficientNet-B7, but a wider and deeper version of the largest model EfficientNet-B7, which is EfficientNet-L2~\cite{efficientnetl2}. All explored convolutional backbones are at resolution 384x384. However, none of them are in the v3.1 release because they do not result in MiDaS models that yield a sufficiently high depth estimation quality.

\subsubsection{Legacy models}

For completeness, we also consider the backbones used in previous MiDaS releases. MiDaS v3.0 is based on the vanilla vision transformer~\cite{vit, steiner2021vit} backbones ViT-L and ViT-B Hybrid at resolution 384x384. It also contains the convolutional encoders of MiDaS v2.1 as legacy backbones, which are ResNeXt-101 32x8d~\cite{resnext-101} at 384x384 (=midas\_v21\_384) and the mobile friendly efficientnet-lite3~\cite{tan2019efficientnet} at 256x256 (=midas\_v21\_256\_small). These four backbones are included as legacy models in MiDaS v3.1. Earlier backbones are not included, which are the convolutional models ResNeXt-101 32x8d~\cite{resnext-101} at 384x384 for MiDaS v2.0 and  ResNet-50~\cite{he2016residual} at 224x224 for MiDaS v1.0. For EfficientNet-Lite3, MiDaS v3.1 also offers an OpenVINO optimized version (=openvino\_midas\_v21\_small\_256).

\subsection{Integration of Backbones into MiDaS}
\label{sec:architecture_encoder_backbones}
In the following we provide technical details on how the new backbones released in MiDaS v3.1 are implemented; these are BEiT\textsubscript{512}-L, BEiT\textsubscript{384}-L, BEiT\textsubscript{384}-B, Swin-L, SwinV2-L, SwinV2-B, SwinV2-T, Next-ViT-L ImageNet-1K-6M and LeViT-224~\cite{beit1, swin, swin2, nextvit, levit}. To minimize the implementation effort, we use the PyTorch Image Models (=timm) repository~\cite{rw2019timm} whenever possible, because this repository offers a common interface to easily exchange backbones. Different backbones are called using a timm function for creating models by providing the name of the desired model. The only exception is Next-ViT, which is not supported by timm but uses it under the hood; we import Next-ViT~\cite{nextvit} as an external dependency.

Since the backbones were trained for image classification they do not inherently contain depth estimation functionality. New encoder backbones used in MiDaS are just feature extractors and need to be connected to the depth decoder appropriately. However, all the new backbones share the common property that they process the input image via successive encoding stages similar to the decoding stages present in the depth decoder. Hence, the task of integrating a new backbone is to properly connect encoding and decoding stages by placing appropriate hooks. This means that we take a tensor computed in the encoder and make it available as input for the decoder at one of its stages. This may require extra operators changing the shape of such tensors to fit to the decoder.

\subsubsection{BEiT}
We begin with the technical details of the BEiT encoder backbones~\cite{beit1}. Getting BEiT transformers instead of the already existing vanilla vision transformers into MiDaS is straightforward, because we can use the timm model creation function mentioned above and use the same hooking mechanism already available in MiDaS v3.0 for ViT~\cite{vit}. We specify the hooks by providing absolute hook positions with respect to the transformer blocks present in the BEiT encoders. Following the hooks chosen for ViT, we select the absolute hook positions 5, 11, 17, 23 for BEiT\textsubscript{512}-L and BEiT\textsubscript{384}-L as well as 2, 5, 8, 11 for BEiT\textsubscript{384}-B. The intuition behind this choice is that the positions are equidistant with one position being at the last transformer block and a gap at the beginning. In addition to that, connecting the encoder backbone also requires a choice of channels for the connected stages, because all transformer blocks of the new encoders contain the same number of channels whereas the depth decoder has different channel numbers per hierarchy level. Here, we also follow the values available for ViT such that we get 256, 512, 1024, 1024 for the number of channels per stage for BEiT\textsubscript{512}-L and BEiT\textsubscript{384}-L as well as 96, 192, 384, 768 for BEiT\textsubscript{384}-B. 
Note that the hook positions and number of channels per stage are based on the MiDaS v3.0 choices and might not be optimal.

There is one important point which makes the implementation of the BEiT transformers in MiDaS v3.1 non-trivial. Although the implementation of BEiT in timm allows arbitrary window sizes, only one such size can be chosen per BEiT encoder created with the timm model creation function. To enable different input resolutions without having to recreate the model, we have modified the original BEiT code of timm by overwriting several timm functions inside of MiDaS. The key problem here is that the variable relative\_position\_indices, which contains relative position indices, is resolution-dependent. The modification generates new indices whenever an unseen resolution is encountered in a single MiDaS run, which may slightly impact performance; for previously encountered resolutions the already computed indices are reused.

\subsubsection{Swin}
Similarly, the Swin and SwinV2 transformers~\cite{swin, swin2} also share the same basic implementation in MiDaS v3.1. A key difference to BEiT and ViT, however, is that Swin and SwinV2 are hierarchical encoders, which changes the structure of the transformer blocks. BeiT and ViT encoders are based on a series of vision transformer blocks whose output is a tensor of rank 2, with always the same shape, where one dimension reflects the number of patches (plus 1 for the class token) and the other one is the embedding dimension. In contrast, for the hierarchical encoders, there are successive hierarchy levels, where each level contains multiple transformer blocks. Going down one hierarchy level halves the resolution in each of the two image directions such that the number of patches increases by 4, whereas the size of the embedding space doubles. The output shape of the transformer blocks is therefore constant only within a hierarchy level but not across them. The advantage of this structure is that we can omit some of the operators, like convolutional and fully connected layers, which are used for ViT and BEiT to change the resolution and number of channels for the hooked tensors of the encoder backbone to fit into the depth decoder. Instead, only transpose and unflatten operators are required.

A consequence of the hierarchical structure is that there has to be exactly one hook per hierarchy level, \ie,~the hooks cannot be chosen freely. The hooks of the Swin and SwinV2 transformers are therefore provided as relative positions with respect to the first transformer block in a hierarchy level. We choose the positions of the hooks as large as possible to reflect the behavior of ViT and BEiT where the last transformer block is always hooked. We thus get the relative hook positions 1, 1, 17, 1 for all three backbones Swin-L, SwinV2-L and SwinV2-B. Note that we did not perform ablations to evaluate how reasonable this choice is. For the number of channels per hierarchy level, we cannot make a choice but we are forced to the numbers provided by the backbones themselves, which are 192, 384, 768, 1536 for Swin-L and SwinV2-L and 128, 256, 512, 1024 for SwinV2-B.

\subsubsection{Next-ViT}
The next encoder type is Next-ViT-L ImageNet-1K-6M~\cite{nextvit}, which is also a hierarchical transformer with 4 stages. Each stage consists of \emph{next} transformer blocks and \emph{next} convolution blocks. Similar to the Swin and SwinV2 transformers, we choose the last block per hierarchy level for the hooks. However, as the implementation of the blocks in Next-ViT is sequential, we do not provide relative hook positions but absolute ones, because this simplifies the implementation. The allowed ranges are 0-2, 3-6, 7-36, 37-39 and we choose the hook positions as 2, 6, 36, 39. The number of channels per hook is again given by the encoder backbone and is this time 96, 256, 512, 1024 (see Table 3 in~\cite{nextvit}). A difference to  Swin and SwinV2 is that the output tensors of the hooked blocks are tensors of rank 3 and not rank 2, where the resolution in the blocks drops from 96x96 to 12x12 for a square input resolution and the number of channels increases from 96 to 1024. Therefore, no extra operators are required to change the shape of these tensors and they can directly be connected to the depth decoder stages. Note that also non-square resolutions are supported. Another important point is that there is a convolutional stem at the beginning of Next-ViT which does already a part of the encoding from the resolution 384x384 down to 96x96. This can be compared to the convolutional patching in front of for example ViT, which also causes a resolution reduction.

\subsubsection{LeViT}
A key difference to the previous backbones is that LeViT \cite{levit}, although also being a hierarchical encoder, is based on only three hierarchy levels. Therefore, we reduce the depth decoder to three hierachy levels for this backbone. To still be able to process images of the resolution 224x224, LeViT-224 utilizes an extra convolutional stem before the attention part, which reduces the resolution to the small value of 14x14. To counter this effect, we insert a similar deconvolutional decoder into the depth decoder. The depth decoder consists of a hierarchical part and a head. The deconvolutional decoder is inserted between these two parts. The convolutional encoder consists of four times the block (Conv2D, BatchNorm2d) with a Hardswish activation function~\cite{howard2019searching} in between each two blocks. For the deconvolutional decoder, we take two (ConvTranspose2D, BatchNorm2d) blocks with Hardswish in between them and also at the end (kernel size 3 and stride 2 as for the convolutional encoder). Only two instead of four blocks are used, because this is sufficient to get the resolution of the depth maps in MiDaS equal to the input resolution with minimal changes to the depth decoder.

We also have to look at the number of channels per processing stage. The four blocks of the encoder stem increase the 3 RGB channels to 16 $\rightarrow$ 32 $\rightarrow$ 64 $\rightarrow$ 128. The depth decoder on the other hand has to decrease the number of channels in multiple likewise processing stages. The hierarchical part of the depth decoder has 256 output channels, which is a fixed number across all backbones of MiDaS v3.1, a choice taken over from MiDaS v3.0. For other backbones, this number is successively decreased to 128 $\rightarrow$ 32 $\rightarrow$ 1, where 1 is the single channel required to represent inverse relative depth. However, for LeViT, the extra deconvolutional decoder already yields a decrease to 128 $\rightarrow$ 64 at the beginning of the depth decoder head. Therefore, the remaining channel reduction has to be adjusted and we use 32 $\rightarrow$ 8 $\rightarrow$ 1 to have a gradual decrease.

For the hooks, the situation is similar to the Swin and SwinV2 transformers, where the tensors hooked in the encoder backbone are of rank 2 such that only transposition and unflattening operators are required to get a shape fitting to the depth decoder. The hook positions are absolute and chosen as 3, 11, 21.

\subsubsection{Others}
The other backbones explored but not released are Next-ViT-L ImageNet-1K, ViT-L Hybrid, vanilla DeiT3-L, DeiT3-L pretrained on ImageNet-22k and fine-tuned on ImageNet-1K, MobileViTv2-0.5, MobileViTv2-2.0, ConvNeXt-L, ConvNeXt-XL and EfficientNet-L2~\cite{nextvit, vit, deit3, mehta2022mobilevitv2, convnext, efficientnetl2}. The first four backbones do not require any new functionality. Next-ViT-L reuses the modifications introduce earlier for Next-ViT-L ImageNet-1K-6M. ViT-L Hybrid is just another variant of ViT-B Hybrid, which is part of MiDaS v3.0. The two DeiT3 backbones are based on the functionality used for ViT. Hence, only MobileViTv2, ConvNeXt and EfficientNet-L require a modification of the MiDaS code. However, this modification is trivial in all these cases, as there are always four hierarchy levels which can directly be hooked into the depth decoder without extra conversion operators.
For MobileViTv2, there is not even a free choice in how the hooks can be chosen. For ConvNeXt and EfficientNet-L, we have proceeded similar to the hooking mechanisms explained earlier. The relative hook positions selected for ConvNeXt are 2, 2, 26, 2, with the allowed ranges 0-2, 0-2, 0-26, 0-2; for EfficientNet-L, this choice is 10, 10, 15, 5, with the ranges 0-10, 0-10, 0-15, 0-5.

\subsection{Training Setup}
\label{sec:training_setup}

We follow the same experimental protocol used in training MiDaS v3.0~\cite{Ranftl_2021_ICCV_DPT} that uses multi-objective optimization~\cite{sener2018opt} with Adam~\cite{adam}, setting the learning rate to 1e-5 for updating the encoder backbones and 1e-4 for the decoder. Encoders are initialized with ImageNet~\cite{deng2009imagenet} weights, whereas decoder weights are initialized randomly. Our training dataset mix is comprised of up to 12 datasets. Similar to~\cite{ranftl2020towards}, we first pretrain models on a subset of the dataset mix for 60 epochs (first training stage), and then train for 60 epochs on the full dataset (second training stage).

\mypara{Dataset Mix 3+10.} This mix is identical to the one used in training MiDaS v3.0. The 10 datasets used include ReDWeb~\cite{xian2018monocular}, DIML~\cite{diml}, Movies~\cite{ranftl2020towards}, MegaDepth~\cite{li2018megadepth}, WSVD~\cite{wang2019web}, TartanAir~\cite{tartanair2020iros}, HRWSI~\cite{xian2020structure}, ApolloScape~\cite{apolloscape}, BlendedMVS~\cite{yao2020blendedmvs}, and IRS~\cite{irs}. A subset consisting of 3 datasets (ReDWeb, HRWSI, BlendedMVS) is used for pretraining models prior to training on the full 10 datasets.

\mypara{Dataset Mix 5+12.} This mix extends the one described above by including NYUDepth v2~\cite{Silberman2012} and KITTI~\cite{Menze_2015_CVPR}. These two datasets were kept out of the training mix in earlier versions of MiDaS to enable zero-shot testing. Our decision to include these two datasets in training is motivated by applications where MiDaS is integrated into metric depth estimation pipelines; we observe that additional training data bolsters model generalizability to indoor and outdoor domains in those applications. In experiments that use this extended dataset mix, a subset now consisting of 5 datasets (ReDWeb, HRWSI, BlendedMVS, NYU Depth v2, KITTI) is used for pretraining models prior to training on the full 12 datasets.

\subsection{Discussion on using New Backbones}
Finally, we describe a general strategy for adding new backbones to the MiDaS architecture for possible future extensions; please refer to \cref{sec:architecture_encoder_backbones} for examples. The main steps are as follows. If possible, the PyTorch Image Models repository~\cite{rw2019timm} or a comparable framework should be used to create a new encoder backbone to reduce the implementation effort. This backbone has to be connected to the depth decoder which requires a choice of hook positions in the encoder backbone. Depending on the shape of the tensors used for the hooking, a series of operators may be required to change the shape such that it fits to the corresponding inputs in the depth decoder. If a backbone contains multiple fundamentally different parts like a convolutional stem at the beginning and an attention part afterwards, the easiest approach is to do the hooking only on the attention part, if possible. To get reasonable resolutions during the depth decoding, it may be required to modify either its hierarchical part or head. This can mean changing the number of hierarchy stages within the network or inverting operators in encoder backbones and inserting them into decoder heads (as we did when integrating the LeViT backbone). Finally, the number of channels at certain network layers may need to be adapted; for this, a helpful guideline may be the structure of similar backbones that have been previously integrated.

\section{Experiments}

In this section, we describe the evaluation protocol and present a comparison of the various models in MiDaS v3.1 alongside a few legacy models from previous releases. We then cover ablation studies that were performed as we experimented with modifying the backbones being incorporated into MiDaS.

\subsection{Evaluation}
\label{sec:evaluation}
Models are evaluated on six datasets: DIW~\cite{chen2016single}, ETH3D~\cite{schops2017multi}, Sintel~\cite{butler2012naturalistic}, KITTI~\cite{Menze_2015_CVPR}, NYU Depth v2~\cite{Silberman2012} and TUM~\cite{sturm2012benchmark}. The type of error computed for each dataset is given by the choice made in the original MiDaS paper~\cite{ranftl2020towards}. For DIW, the computed metric is the Weighted Human Disagreement Rate (WHDR). For ETH3D and Sintel, the mean absolute value of the relative error (REL) $\frac{1}{M}\sum_{i=1}^M\left|d_i-d_i^*\right|/d_i^*$ is used, where M is the number of pixels, $d_i$ is the relative depth and the asterisk, \eg, $d_i^*$, denotes the ground truth. For the remaining three datasets, the percentage of bad depth pixels $\delta_1$ with $\max(d_i/d_i^*,d_i^*/d_i)>1.25$ is counted.

\begin{figure*}
    \centering
    \captionsetup{type=figure}
    \includegraphics[width=\linewidth]{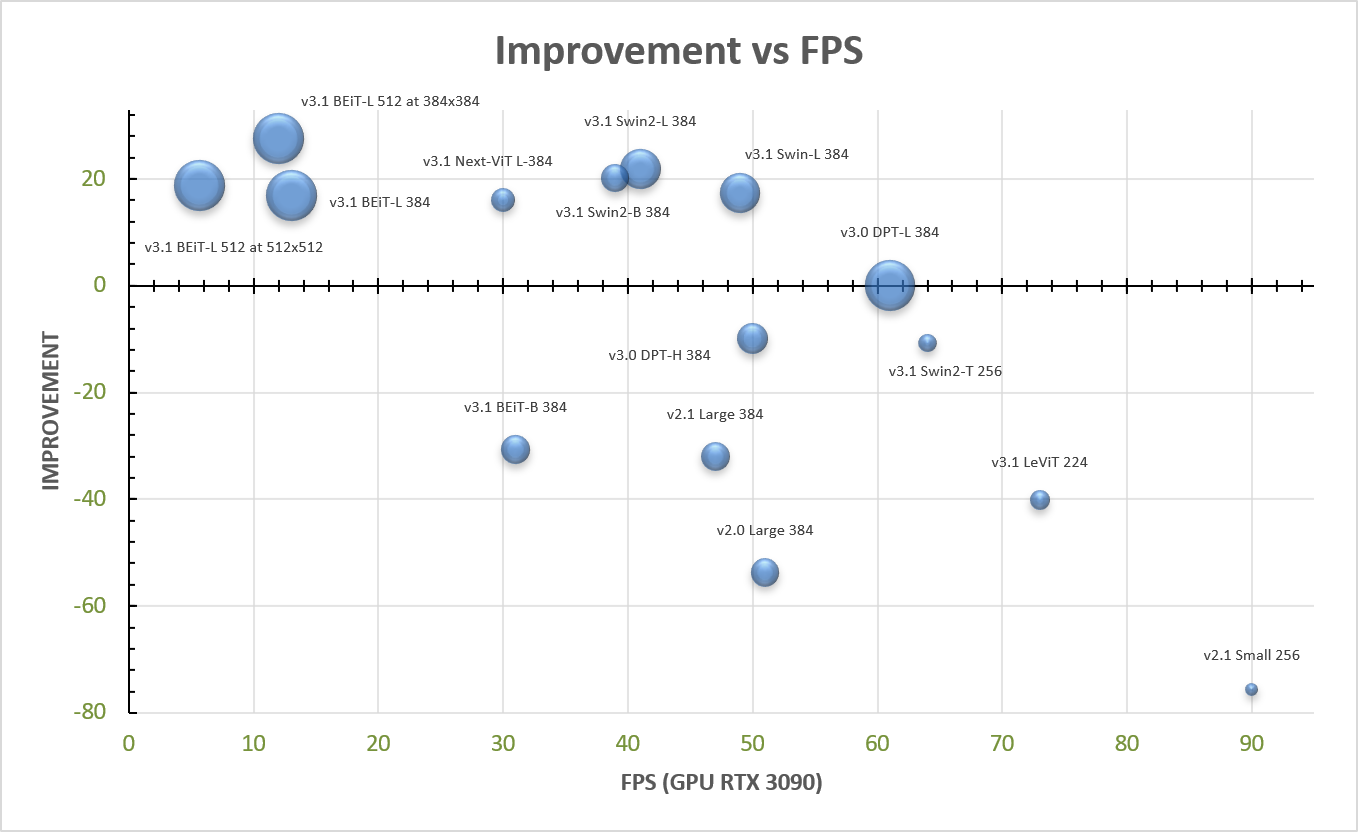}
    \captionof{figure}{\textbf{Improvement vs FPS.} The plot shows the improvement of all the models of MiDaS v3.1 with respect to the largest model DPT\textsubscript{L 384} (=ViT-L 384) of MiDaS v3.0 vs the frames per second. The framerate is measured on an RTX 3090 GPU. The area covered by the bubbles is proportional to the number of parameters of the corresponding models. In the model descriptions, we provide the MiDaS version, because some models of MiDaS v3.1 are legacy models which were already introduced in earlier MiDaS releases. The first 3-digit number in the model name reflects the training resolution which is always a square resolution. For two BEiT models, we also provide the inference resolution at the end of the model description, because there the inference resolution differs from the training one. The improvement is defined as the relative zero-shot error averaged over six datasets as explained in \cref{sec:evaluation}.}
    \label{fig:improvement_vs_fps}
\end{figure*}

For a quick model comparison, we introduce the relative improvement with respect to the largest model ViT-L 384 from MiDaS v3.0. The relative improvement is defined as the relative zero-shot error averaged over the six datasets. Denoting all the errors as $\epsilon_s$, with $s\in\{1,...,6\}$ being the dataset index, the improvement is then defined as
\begin{equation}\label{eq:improvement}
I=100\left(1-\frac{1}{6}\sum_d\frac{\epsilon_d}{\epsilon_{d,\rm{ViT-L 384}}}\right)\%
\end{equation}
where $\epsilon_{d,\rm{ViT-L 384}}$ are the respective errors for the model ViT-L 384. Note that a difference in resolution limits the comparability of the zero-shot errors and thus the improvement. This is because these quantities are averages over the pixels of an image and do not take into account the potential advantage of more details present at higher resolutions. A visualization of the relative improvement versus the frame rate is shown in in \cref{fig:improvement_vs_fps}.

We also use the root mean square error of the disparity (RMSE) $[\frac{1}{M}\sum_{i=1}^M\left|D_i-D_i^*\right|^2]^{\frac{1}{2}}$, where $D_i$ is the disparity, for additional comparisons of models during training (cf.~\cref{tab:model_overview_1st_stage}).

\subsection{Results and Analysis}
\label{sec:results_and_analysis}
An overview of the validation results is provided in Tabs.~\ref{tab:model_overview_2nd_stage_released}, \ref{tab:model_overview_2nd_stage_unreleased}  and \ref{tab:model_overview_1st_stage}. While Tabs.~\ref{tab:model_overview_2nd_stage_released} and \ref{tab:model_overview_2nd_stage_unreleased} show completely trained models, \ie,~training is done in two stages, the models in~\cref{tab:model_overview_1st_stage} are not trained beyond the first stage (cf.~\cref{sec:training_setup}) since the depth estimation quality observed there is too low to justify further training. These models are presented despite incomplete training to show both accepted and discarded backbones. In this section, we discuss the models in~\cref{tab:model_overview_2nd_stage_released}, those above the horizontal separator in~\cref{tab:model_overview_2nd_stage_unreleased} and the models between the first and last horizontal separators of~\cref{tab:model_overview_1st_stage}. The remaining models are either included for comparisons or they are experimental. A thorough explanation of them can be found in~\cref{sec:ablation_studies}.

\subsubsection{Published Models}
\cref{tab:model_overview_2nd_stage_released} contains the models released as a part of MiDaS v3.1. BEiT\textsubscript{512}-L is the best model for both square and unconstrained resolutions. Note that unconstrained resolutions mean an aspect ratio defined by the dataset. The quality of the BEiT\textsubscript{512}-L model can be seen from the relative improvement $I$ in~\cref{tab:model_overview_2nd_stage_released}, which is 36\% for square resolutions and 19\% for resolutions of height 512 as well as 28\% if the height is 384. Note that different inference resolutions have to be considered separately here due to the limitations of the relative improvement $I$ mentioned in~\cref{sec:evaluation}.

MiDaS v3.1 includes more models than earlier versions to provide a better coverage of possible downstream tasks, including lightweight models. This is reflected by new models like LeViT-224 in~\cref{tab:model_overview_2nd_stage_released}, which is the fastest new model with a framerate of 73 frames per second (fps). It is surpassed in speed only by the legacy model EfficientNet-Lite3 that runs at 90 fps.

\subsubsection{Unpublished Models}
The models in~\cref{tab:model_overview_2nd_stage_unreleased} are not released due to a lower depth estimation quality compared to the released ones. The first of these models is Swin-L, trained on the dataset configuration 3+10. Here, we have released only the variant trained on the configuration 5+12, as shown in~\cref{tab:model_overview_2nd_stage_released}. As we see from the rightmost column of Tabs.~\ref{tab:model_overview_2nd_stage_released} and \ref{tab:model_overview_2nd_stage_unreleased}, the increased number of datasets improves the quality measure $I$ from 2\% to 21\%, which is a significant jump. The main contribution for this increase comes from KITTI and NYUDepth v2 no longer being zero-shot datasets when trained with the configuration 5+12. This can be seen from the decrease of the $\delta_1$ scores of KITTI and NYUDepth v2 from 12.15 and 6.571 to 6.601 and 3.343 respectively, while the remaining errors decrease only slightly (see Tabs.~\ref{tab:model_overview_2nd_stage_released} and \ref{tab:model_overview_2nd_stage_unreleased}). The next unreleased model in~\cref{tab:model_overview_2nd_stage_unreleased} is Swin-T, which is not part of MiDaS v3.1, because SwinV2 generally yields better results than Swin. Finally, we have also studied the MobileViTv2 family of transformers, which contains MobileViTv2-0.5 as our smallest model with 13 million parameters. However, both variants MobileViTv2-0.5 and MobileViTv2-2.0 have values of $I$ around -300\%, which reflects a too low quality to be relevant.

As the models below the horizontal separator of~\cref{tab:model_overview_2nd_stage_unreleased} are explained in~\cref{sec:ablation_studies}, we proceed with the models between the first and last horizontal separator of~\cref{tab:model_overview_1st_stage}. The models shown there split into models with transformer and convolutional encoder backbones, which are separated by the dashed separator. We start with the transformer models, where we first have DeiT3-L-22K-1K and DeiT3-L. These two models have a high depth estimation quality, \eg, 0.070 for the relative error (REL) of the BlendedMVS dataset, which is equal to the value of BEiT\textsubscript{384}-L also visible in~\cref{tab:model_overview_2nd_stage_unreleased} for a comparison. However, as the DeiT3 transformers do not surpass the quality of BEiT\textsubscript{384}-L, we did not train them beyond the first stage. The same criterion holds for ViT-L Hybrid, which was explored, because ViT-B Hybrid is part of MiDaS v3.0 (cf.~\cref{tab:model_overview_2nd_stage_released}). For Next-ViT-L-1K and Next-ViT-L-1K-6M, we have decided to include the better of the two variants in MiDaS v3.1, which is Next-ViT-L-1K-6M according to~\cref{tab:model_overview_1st_stage}. 

Finally, we have also explored the three convolutional models ConvNeXt-XL, ConvNeXt-L and EfficientNet-L2. As we explored them with the intention to get a model of highest quality and it did not beat BEiT\textsubscript{384}-L, we have discarded these models. In particular, EfficientNet-L2 shows a low depth estimation quality with errors of 0.165, 0.227 and 0.219 according to~\cref{tab:model_overview_1st_stage}.

\begin{table*}[!htb]
\footnotesize
\centering
\setlength{\tabcolsep}{2pt} % cols space (default: 6pt)
\begin{tabular}
{@{}
    l@{\hspace{6pt}}c@{\hspace{3pt}}|
    c@{\hspace{6pt}}c@{\hspace{3pt}}|
    c@{\hspace{3pt}}c@{\hspace{3pt}}c@{\hspace{3pt}}c@{\hspace{3pt}}c@{\hspace{3pt}}c@{\hspace{3pt}}|c@{\hspace{3pt}}|
    c@{\hspace{3pt}}c@{\hspace{3pt}}c@{\hspace{3pt}}c@{\hspace{3pt}}c@{\hspace{3pt}}c@{\hspace{3pt}}|c@{\hspace{3pt}}
    @{}}
    
\toprule

\multicolumn{2}{c|}{\textbf{Model}} & \multicolumn{2}{c|}{\textbf{Resources}} & \multicolumn{7}{c|}{\textbf{Unconstrained Resolution}} & \multicolumn{7}{c}{\textbf{Square Resolution}} \\

                                                       & Data & Par.           & FPS            & DIW                & ETH3D             & Sintel            & KITTI                  & NYU                    & TUM                    & I              & DIW                & ETH3D             & Sintel            & KITTI                  & NYU                    & TUM                    & I \\ 

Encoder/Backbone                                       & Mix  & \,$\downarrow$ & $\uparrow$     & WHDR\,$\downarrow$ & REL\,$\downarrow$ & REL\,$\downarrow$ & $\delta_1\,\downarrow$ & $\delta_1\,\downarrow$ & $\delta_1\,\downarrow$ & \%\,$\uparrow$ & WHDR\,$\downarrow$ & REL\,$\downarrow$ & REL\,$\downarrow$ & $\delta_1\,\downarrow$ & $\delta_1\,\downarrow$ & $\delta_1\,\downarrow$ & \%\,$\uparrow$ \\ 

\midrule

BEiT\textsubscript{512}-L~\cite{beit1}                 & 5+12 & 345            & 5.7            & 0.114              & \textbf{0.066}    & 0.237             & 11.57*                 & \textbf{1.862}*        & \textbf{6.132}         & \underline{19} & 0.112              & \textbf{0.061}    & \textbf{0.209}    & \textbf{5.005}*        & \textbf{1.902}*        & \textbf{6.465}         & \textbf{36} \\
BEiT\textsubscript{384}-L~\cite{beit1}                 & 5+12 & 344            & 13             & 0.124              & \underline{0.067} & 0.255             & 9.847*                 & 2.212*                 & \underline{7.176}      & 16.8           & 0.111              & \underline{0.064} & \underline{0.222} & \underline{5.110}*     & \underline{2.229}*     & 7.453                  & \underline{33.0} \\
BEiT\textsubscript{512}-L@384~\cite{beit1}             & 5+12 & 345            & 5.7            & 0.125              & 0.068             & \textbf{0.218}    & \textbf{6.283}*        & \underline{2.161}*     & \textbf{6.132}         & \textbf{28}    & 0.117              & 0.070             & 0.223             & 6.545*                 & 2.582*                 & \underline{6.804}      & 29 \\
SwinV2-L\cite{swin2}                                   & 5+12 & 213            & 41             & --                 & --                & --                & --                     & --                     & --                     & --             & 0.111              & 0.073             & 0.244             & 5.840*                 & 2.929*                 & 8.876                  & 25 \\
SwinV2-B\cite{swin2}                                   & 5+12 & 102            & 39             & --                 & --                & --                & --                     & --                     & --                     & --             & \underline{0.110}  & 0.079             & 0.240             & 5.976*                 & 3.284*                 & 8.933                  & 23 \\
Swin-L\cite{swin}                                      & 5+12 & 213            & 49             & --                 & --                & --                & --                     & --                     & --                     & --             & 0.113              & 0.085             & 0.243             & 6.601*                 & 3.343*                 & 8.750                  & 21 \\
BEiT\textsubscript{384}-B~\cite{beit1}                 & 5+12 & 112            & 31             & 0.116              & 0.097             & 0.290             & 26.60*                 & 3.919*                 & 9.884                  & -31            & 0.114              & 0.085             & 0.250             & 8.180*                 & 3.588*                 & 9.276                  & 16 \\
Next-ViT-L-1K-6M\cite{nextvit}                         & 5+12 & 72             & 30             & \textbf{0.103}     & 0.095             & \underline{0.230} & \underline{6.895}*     & 3.479*                 & 9.215                  & 16             & \textbf{0.106}     & 0.093             & 0.254             & 8.842*                 & 3.442*                 & 9.831                  & 14 \\
\textit{ViT-L}~\cite{vit}                              & 3+10 & 344            & 61             & \underline{0.108}  & 0.089             & 0.270             & 8.461                  & 8.318                  & 9.966                  & 0              & 0.112              & 0.091             & 0.286             & 9.173                  & 8.557                  & 10.16                  & 0 \\
\textit{ViT-B Hybrid}~\cite{vit}                       & 3+10 & 123            & 61             & 0.110              & 0.093             & 0.274             & 11.56                  & 8.69                   & 10.89                  & -10            & -                  & -                 & -                 & -                      & -                      & -                      & - \\
SwinV2-T\cite{swin2}                                   & 5+12 & \underline{42} & 64             & --                 & --                & --                & --                     & --                     & --                     & --             & 0.121              & 0.111             & 0.287             & 10.13*                 & 5.553*                 & 13.43                  & -6 \\
\textit{ResNeXt-101}~\cite{resnext-101}                & 3+10 & 105            & 47             & 0.130              & 0.116	         & 0.329             & 16.08                  & 8.71	               & 12.51                  & -32            & -                  & -                 & -                 & -                      & -                      & -                      & - \\
LeViT-224\cite{levit}                                  & 5+12 & 51             & \underline{73} & --                 & --                & --                & --                     & --                     & --                     & --             & 0.131              & 0.121             & 0.315             & 15.27*                 & 8.642*                 & 18.21                  & -34 \\
\textit{EfficientNet-Lite3}~\cite{tan2019efficientnet} & 3+10 & \textbf{21}    & \textbf{90}    & 0.134	             & 0.134             & 0.337             & 29.27	              & 13.43                  & 14.53	                & -75            & -                  & -                 & -                 & -                      & -                      & -                      & - \\

\bottomrule
\end{tabular}
\vspace{-6pt}
\caption{\textbf{Evaluation of released models (post second training stage).} The table shows the validation of the second training stage (see~\cref{sec:training_setup}) for the models released in MiDaS v3.1. The dataset definitions 3+10 and 5+12 used for the training can be found in~\cref{sec:training_setup}. The resources required per model are given by the number of parameters in million (Par.)~and the frames per second (FPS, if possible for the unconstrained resolution). The validation is done on the datasets DIW~\cite{chen2016single}, ETH3D~\cite{schops2017multi}, Sintel~\cite{butler2012naturalistic}, KITTI~\cite{Menze_2015_CVPR}, NYU Depth v2~\cite{Silberman2012} and TUM~\cite{sturm2012benchmark} with the validation errors as described in~\cref{sec:evaluation}. The resolution is either unconstrained, \ie~the aspect ratio is given by the images in the dataset, or the images are converted to a square resolution. Overall model quality is given by the relative improvement I with respect to ViT-L (cf.~\cref{eq:improvement}). Note that Next-ViT-L-1K-6M and ResNeXt-101 are short forms of Next-ViT-L ImageNet-1K-6M and ResNeXt-101 32x8d. The suffix @384 means that the model is validated at the inference resolution 384x384 (differing from the training resolution). Legacy models from MiDaS v3.0 and 2.1 are in italics, where ResNeXt-101=midas\_v21\_384 and Efficientnet-lite3=midas\_v21\_256\_small. Validation errors that could not be evaluated, because of the model not supporting the respective resolution are marked by --. Quantities not evaluated due to other reasons are given by -. The asterisk * refers to non-zero-shot errors, because of the training on KITTI and NYU Depth v2. The rows are ordered such that models with better relative improvement values for the square resolution are at the top. The best numbers per column are bold and second best underlined.}
\label{tab:model_overview_2nd_stage_released}
\end{table*}

\begin{table*}[!htb]
\footnotesize
\centering
\setlength{\tabcolsep}{2pt} % cols space (default: 6pt)
\begin{tabular}
{@{}
    l@{\hspace{6pt}}c@{\hspace{3pt}}|
    c@{\hspace{6pt}}c@{\hspace{3pt}}|
    c@{\hspace{3pt}}c@{\hspace{3pt}}c@{\hspace{3pt}}c@{\hspace{3pt}}c@{\hspace{3pt}}c@{\hspace{3pt}}|c@{\hspace{3pt}}|
    c@{\hspace{3pt}}c@{\hspace{3pt}}c@{\hspace{3pt}}c@{\hspace{3pt}}c@{\hspace{3pt}}c@{\hspace{3pt}}|c@{\hspace{3pt}}
    @{}}
    
\toprule

\multicolumn{2}{c|}{\textbf{Model}} & \multicolumn{2}{c|}{\textbf{Resources}} & \multicolumn{7}{c|}{\textbf{Unconstrained Resolution}} & \multicolumn{7}{c}{\textbf{Square Resolution}} \\

                                            & Data     & Par.           & FPS            & DIW                & ETH3D             & Sintel            & KITTI                  & NYU                    & TUM                    & I              & DIW                & ETH3D             & Sintel            & KITTI                  & NYU                    & TUM                    & I \\ 

Encoder/Backbone                            & Mix      & \,$\downarrow$ & $\uparrow$     & WHDR\,$\downarrow$ & REL\,$\downarrow$ & REL\,$\downarrow$ & $\delta_1\,\downarrow$ & $\delta_1\,\downarrow$ & $\delta_1\,\downarrow$ & \%\,$\uparrow$ & WHDR\,$\downarrow$ & REL\,$\downarrow$ & REL\,$\downarrow$ & $\delta_1\,\downarrow$ & $\delta_1\,\downarrow$ & $\delta_1\,\downarrow$ & \%\,$\uparrow$ \\ 

\midrule

Swin-L\cite{swin2}                          & 3+10     & 213            & 41             & --                 & --                & --                & --                     & --                     & --                     & --             & 0.115              & 0.086             & 0.246             & 12.15                  & 6.571                  & 9.745                  & 2 \\
Swin-T\cite{swin2}                          & 3+10     & 42             & \underline{71} & --                 & --                & --                & --                     & --                     & --                     & --             & 0.131              & 0.120             & 0.334             & 15.66                  & 12.69                  & 14.56                  & -38 \\
MobileViTv2-0.5~\cite{mehta2022mobilevitv2} & 5+12     & \textbf{13}    & \textbf{72}    & 0.430              & 0.268             & 0.418             & 51.77*                 & 45.32*                 & 39.33                  & -301           & 0.509              & 0.263             & 0.422             & 37.67*                 & 48.65*                 & 40.63                  & -286 \\
MobileViTv2-2.0~\cite{mehta2022mobilevitv2} & 5+12     & \underline{34} & 61             & 0.509              & 0.263             & 0.422             & 37.67*                 & 48.65*                 & 40.63                  & -294           & 0.501              & 0.269             & 0.433             & 59.94*                 & 48.32*                 & 41.79                  & -320 \\

\midrule

BEiT\textsubscript{384}-L 5K+12K            & $\cdot$K & 344            & 13             & 0.120              & 0.066             & \underline{0.213} & \underline{2.967}*     & 2.235*                 & \textbf{6.570}         & \underline{35} & \underline{0.110}  & \underline{0.066} & \textbf{0.212}    & 5.929*                 & 2.296*                 & \textbf{6.772}         & \textbf{33} \\
BEiT\textsubscript{384}-L Wide              & 5+12     & 344            & 13             & \underline{0.111}  & 0.068             & 0.247             & 10.73*                 & 2.146*                 & \underline{7.217}      & 17.4           & 0.112              & \underline{0.066} & 0.221             & \textbf{5.078}*        & \underline{2.216}*     & \underline{7.401}      & \textbf{32.7} \\
BEiT\textsubscript{384}-L 5+12+12K          & +12K     & 344            & 13             & 0.123              & \underline{0.065} & 0.216             & \underline{2.967}*     & \underline{2.066}*     & 7.417                  & 33             & \textbf{0.107}     & \textbf{0.064}    & 0.217             & \underline{5.631}*     & 2.259*                 & 7.659                  & \underline{32} \\
BEiT\textsubscript{384}-L A5+12A            & $\cdot$A & 344            & 13             & \textbf{0.110}     & \textbf{0.061}    & \textbf{0.207}    & \textbf{2.802}*        & \textbf{1.891}*        & 7.533                  & \textbf{37}    & 0.113              & 0.070             & \underline{0.213} & 6.504*                 & \textbf{2.179}*        & 7.946                  & 29 \\

\bottomrule
\end{tabular}
\vspace{-6pt}
\caption{\textbf{Evaluation of unpublished models (post second training stage).} The table shows the validation of the second training stage (see~\cref{sec:training_setup}) of models not released in MiDaS v3.1 due to a low depth estimation quality. The models below the horizontal separator are based on experimental modifications explained in~\cref{sec:ablation_studies}. The general table layout is similar to~\cref{tab:model_overview_2nd_stage_released}. The extra dataset mixes, like $\cdot$K, are explained in~\cref{sec:ablation_studies}.}
\label{tab:model_overview_2nd_stage_unreleased}
\end{table*}

\begin{table}[!htb]
\footnotesize
\centering
\setlength{\tabcolsep}{2pt} % cols space (default: 6pt)
\begin{tabular}
{@{}
    l@{\hspace{6pt}}|
    c@{\hspace{6pt}}c@{\hspace{3pt}}c@{\hspace{3pt}}
    @{}}
    
\toprule

& \multicolumn{3}{c}{\textbf{Square Resolution}} \\

                                                                     & HRWSI              & BlendedMVS         & ReDWeb \\ 

Model                                                                & RMSE\,$\downarrow$ & REL\,$\downarrow$  & RMSE\,$\downarrow$ \\ 

\midrule

BEiT\textsubscript{384}-L~\cite{beit1}                               & \textbf{0.068}     & \textbf{0.070}     & \textbf{0.076} \\
Swin-L\cite{swin} Training 1                                         & 0.0708             & \underline{0.0724} & 0.0826 \\
\phantom{Swin-L\cite{swin}} Training 2                               & 0.0713             & \underline{0.0720} & 0.0831 \\
\textit{ViT-L}~\cite{vit}                                            & 0.071              & \underline{0.072}  & 0.082 \\
Next-ViT-L-1K-6M~\cite{nextvit}                                      & 0.075              & 0.073              & 0.085 \\

\midrule

DeiT3-L-22K-1K~\cite{deit3}                                          & \underline{0.070}  & \textbf{0.070}     & \underline{0.080} \\
ViT-L Hybrid~\cite{vit}                                              & 0.075              & 0.075              & 0.085 \\
Next-ViT-L-1K~\cite{nextvit}                                         & 0.078              & 0.075              & 0.087 \\
DeiT3-L~\cite{deit3}                                                 & 0.077              & 0.075              & 0.087 \\

\hspace{-0.001cm}\hdashrule[0.5ex]{2.9cm}{0.5pt}{1mm}\hspace{-0.3cm} & \multicolumn{3}{c}{\hdashrule[0.5ex]{4cm}{0.5pt}{1mm}} \\

ConvNeXt-XL~\cite{convnext}                                           & 0.075             & 0.075              & 0.085 \\
ConvNeXt-L~\cite{convnext}                                            & 0.076             & 0.076              & 0.087 \\
EfficientNet-L2~\cite{efficientnetl2}                                 & 0.165             & 0.227              & 0.219 \\

\hspace{-0.001cm}\hdashrule[0.5ex]{2.75cm}{0.5pt}{1pt}\hspace{-0.3cm} & \multicolumn{3}{c}{\hdashrule[0.5ex]{4cm}{0.5pt}{1pt}} \\

ViT-L Reversed                                                        & 0.071             & 0.073              & 0.081 \\
Swin-L Equidistant                                                    & 0.072             & 0.074              & 0.083 \\

%X~\cite{X} & & & & & & \\

\bottomrule
\end{tabular}
\vspace{-6pt}
\caption{\textbf{Model evaluation (post first training stage).} The table shows the validation of unpublished models which were mostly trained only in the first training stage and not also the second one due to low depth estimation quality (see~\cref{sec:training_setup}). The models above the horizontal separator line (between Next-ViT-L-1K-6M and DeiT3-L-22K-1K) are included for a comparison with the other models and have at least a released variant in ~\cref{tab:model_overview_2nd_stage_released}, although they were also not released directly (see~\cref{sec:results_and_analysis} for details). For Swin-L, two different training runs are shown. The models above the dashed separator are models based on transformer backbones, and the models between the dashed and dotted line are convolutional ones. The rows below the dotted separator are models with experimental modifications as explained in~\cref{sec:ablation_studies}. All the models in this table are trained on the 3+10 dataset configuration (in contrast to the mixtures of Tabs.~\ref{tab:model_overview_2nd_stage_released} and \ref{tab:model_overview_2nd_stage_unreleased}). Validation is done on the datasets HRWSI~\cite{xian2020structure}, BlendedMVS~\cite{yao2020blendedmvs} and ReDWeb~\cite{xian2018monocular}. The errors used for validation are the root mean square error of the disparity (RMSE) and the mean absolute value of the relative error (REL), see~\cref{sec:evaluation}. Note that DeiT3-L-22K-1K is DeiT3-L pretrained on ImageNet-22k and fine-tuned on ImageNet-1K, Next-ViT-L-1K is the shortened form of Next-ViT-L ImageNet-1K and Next-ViT-L-1K-6M stands for Next-ViT-L ImageNet-1K-6M. The model in italics is a retrained legacy model from MiDaS v3.0. The rows are ordered such that better models are at the top. The best numbers per column are bold and second best underlined.}
\label{tab:model_overview_1st_stage}
\end{table}

\begin{figure*}
    \centering
    \captionsetup{type=figure}
    \includegraphics[width=\linewidth]{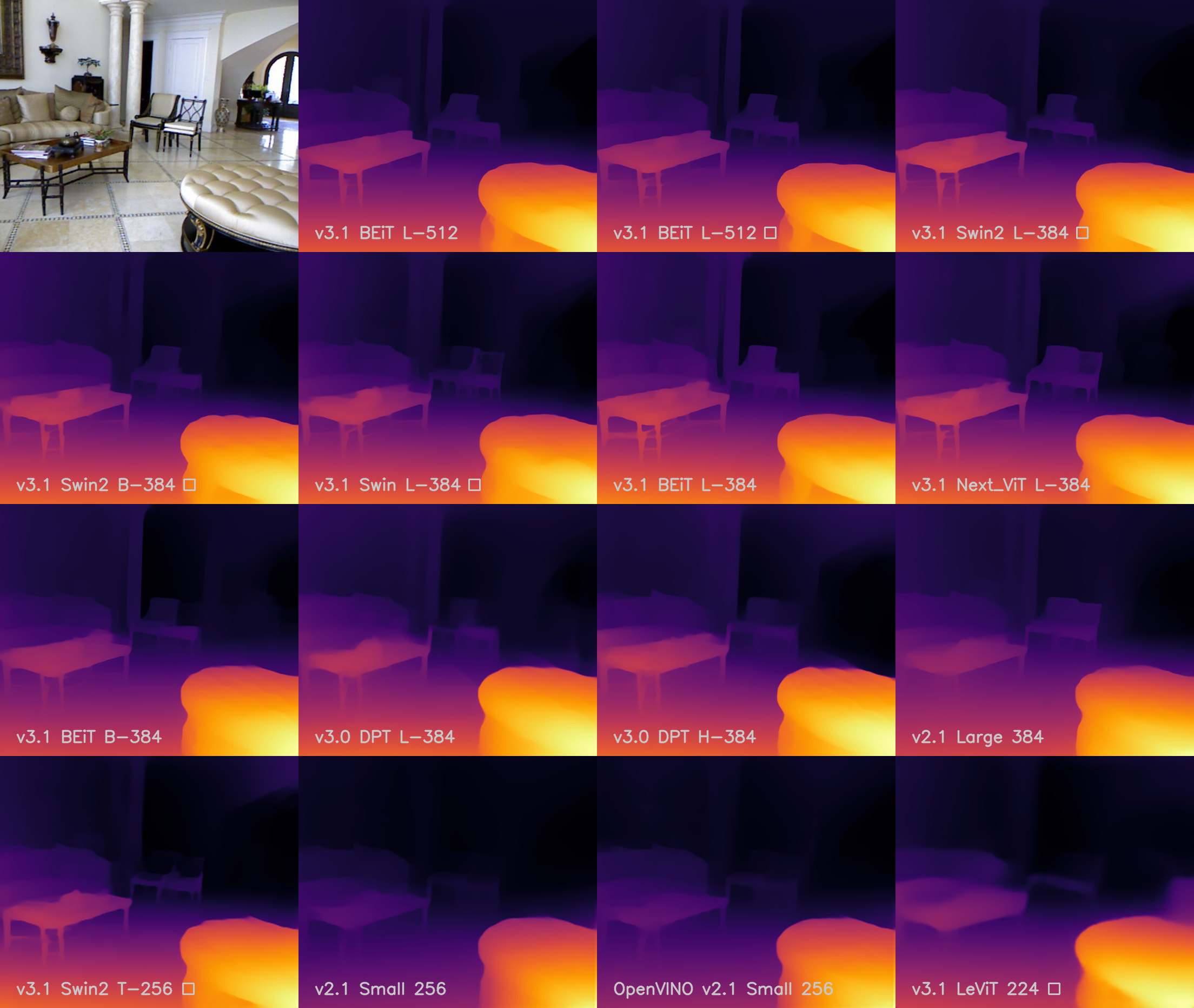}
    \captionof{figure}{\textbf{Backbone comparison.} The table shows the inverse relative depth maps of the different models of MiDaS v3.1, including legacy models, for the example RGB input image at the top, left. The brighter the colors, the larger the inverse relative depths, \ie,~the closer the represented objects are to the camera. The names of the models are shown at the bottom left part of each depth map. This includes the MiDaS version, the backbone name and size as well as the training resolution. Models which are evaluated only at a square resolution are marked by the square symbol at the end of the white texts. The second last model at the bottom row is an OpenVINO model.}
    \label{fig:comparison}
\end{figure*}

\subsection{Ablation Studies}
\label{sec:ablation_studies}

In the following, we discuss experimental modifications of some of the investigated backbones, which helps to get a better understanding of the associated configurations. The modifications can be found at the bottom of~Tabs.~\ref{tab:model_overview_2nd_stage_unreleased} and \ref{tab:model_overview_1st_stage}. In addition to that, we also walk through the models at the top of~\cref{tab:model_overview_2nd_stage_unreleased}, which are included for a comparison with the other models in that table.

We begin with the four reference models at the top of~\cref{tab:model_overview_1st_stage}. Variants of these models are also available in~\cref{tab:model_overview_2nd_stage_released}. For BEiT\textsubscript{384}-L and Next-ViT-L-1K-6M, these are models with different training datasets, \ie.~3+10 in~\cref{tab:model_overview_1st_stage} and 5+12 in~\cref{tab:model_overview_2nd_stage_released}. For Swin-L, no such difference is given between the two tables. However, in~\cref{tab:model_overview_1st_stage}, we have included two separate training runs to provide an approximation of the variance in the training process. ViT-L is basically the same model in both tables, but the training runs are independent, because a retraining was required to get the data required for ~\cref{tab:model_overview_1st_stage}.

We continue with the two experimental modifications at the bottom of~\cref{tab:model_overview_1st_stage}, which have undergone only one training stage. The first modification, denoted as ViT-L Reversed, is the vanilla vision transformer backbone ViT-L already released in MiDaS v3.0, but with the order of the hooks reversed. Instead of providing the depth decoder hooks with the absolute positions 5, 11, 17, 23, we set them to 23, 17, 11, 5. This is possible, because the ViT encoder family is based on a series of similar transformer blocks, which do not differ like the transformer blocks in for instance the hierarchical structure of the Swin transformers. Astonishingly, as shown in~\cref{tab:model_overview_1st_stage}, the reversal of the hooks has practically no impact on the depth estimation quality. So, there is no major difference if the four hierarchy levels of the decoder are connected in forward or reverse order to the transformer blocks of the encoder.

The second experiment is Swin-L Equidistant where the hooks are chosen as equidistantly as possible, similar to ViT-L. As we consider a Swin transformer here, the hook positions are relative and constrained to 0-1, 0-1, 0-17, 0-1 (cf.~\cref{sec:architecture_encoder_backbones}). To homogenize the distance between the hooks, we replace the positions 1, 1, 17, 1 of Swin-L by 1, 1, 9, 1. Note that the distances could be made even more similar by setting the first hook to zero. However, here we follow ViT-L, where a gap is chosen before the first hook. As we see from~\cref{tab:model_overview_1st_stage}, the modification leads to a small decrease of the depth estimation quality when compared to the unmodified model Swin-L such that we have not released the corresponding model. To also get at least a very rough estimate of the significance of this change, we have actually included two independent training runs for Swin-L, denoted by training 1 and 2 in~\cref{tab:model_overview_1st_stage}. As we see, the training variance seems to be rather small for Swin-L.

\cref{tab:model_overview_2nd_stage_unreleased} shows four additional modifications, where we have also trained the second stage. We first consider the model BEiT\textsubscript{384}-L Wide, where the hooks are widened by removing the hook gap at the beginning of the encoder. Instead of the absolute hook positions 5, 11, 17, 23 of BEiT\textsubscript{384}-L in~\cref{tab:model_overview_2nd_stage_released} (see~\cref{sec:architecture_encoder_backbones}), the modification uses 0, 7, 15, 23. As we see from~\cref{tab:model_overview_2nd_stage_unreleased}, there is nearly no impact on the depth estimation quality. For unconstrained resolutions, the relative improvement $I$ is 17.4\% for the widened variant and thus a bit better than the value 16.8\% for the original variant in~\cref{tab:model_overview_2nd_stage_released}. For square resolutions, the situation is the opposite, where we have the values 32.7\% and 33.0\%. With the effect being so small, we have decided to keep the hook gap.

The remaining three modifications in~\cref{tab:model_overview_2nd_stage_unreleased}, denoted as BEiT\textsubscript{384}-L 5+12+12K, BEiT\textsubscript{384}-L 5K+12K and BEiT\textsubscript{384}-L 5A+12A, address the large value $\delta_1=9.847$ of KITTI for the unconstrained resolution of BEiT\textsubscript{384}-L when compared to $\delta_1=2.212$ of NYU Depth v2 in~\cref{tab:model_overview_2nd_stage_released}. The reason for the large $\delta_1$ value is that the training images of KITTI have a high aspect ratio caused by the resolution 1280x384, where the width is much bigger than the height. This is different for \eg, NYU Depth v2, where the resolution is 512x384 and thus the aspect ratio is significantly lower. However, in BEiT\textsubscript{384}-L, the resolution 1280x384 is reduced to 384x384 by random cropping such that there is a strong resolution discrepancy between training and inference, because for the unconstrained resolution inference is done with the original resolution 1280x384. In the modifications, we remove this discrepancy by training KITTI on the original resolution 1280x384. Whenever KITTI is trained in this way, we add the letter K as a suffix after the dataset counter. This leads us to the first modification BEiT\textsubscript{384}-L 5+12+12K, where we take the original model BEiT\textsubscript{384}-L trained in two stages on the data 5+12 and add a third stage, which is also trained on the 12 datasets of the second stage but now with the original KITTI resolution. As we see from \cref{tab:model_overview_2nd_stage_unreleased}, this lowers the $\delta_1$ value from 9.847 to 2.967. Note that for simplicity we only provide the dataset change +12K and not the whole description 5+12+12K in the data column of \cref{tab:model_overview_2nd_stage_released}.

For BEiT\textsubscript{384}-L 5K+12K, we use only two training stages and train them with the original KITTI resolution. Hence, we denote the dataset as 5K+12K instead of 5+12, or $\cdot$K in short. This does not change the $\delta_1$ value of KITTI for the unconstrained resolution, but improves the overall model quality a bit. The relative improvement $I$ increases from 33\% to 35\% for the unconstrained resolution and 32\% to 33\% for the square one. We also test extending the approach to use the original aspect ratio of the training images during training for the other datasets. If the training resolution is not constant over the training images, we use the average resolution, adjusted to a multiple of 32\%. This gives 480x448 for ReDWeb~\cite{xian2018monocular}, 480x448 for MegaDepth~\cite{li2018megadepth}, 384x384 for WSVD~\cite{wang2019web} and 544x384 for HRWSI~\cite{xian2020structure}. The resulting modified model is BEiT\textsubscript{384}-L 5A+12A, where the letter A, standing for `all`, denotes that now all training datasets of the respective stage have a resolution close to the original one ($\cdot$A in the data column of \cref{tab:model_overview_2nd_stage_unreleased}). The consequence of this change is that the $\delta_1$ score of KITTI for the unconstrained resolution drops to the lowest and thus best value 2.802. Also, the relative improvement is best for the modified model, where $I=37\%$. However, there might be an overfitting to the resolution of the training images, because for square resolutions the relative improvement drops from 33\% to 29\% and is thus even below the 36\% of the BEiT\textsubscript{512}-L model of ~\cref{tab:model_overview_2nd_stage_released}. Therefore, we have not released BEiT\textsubscript{384}-L 5A+12A, but it shows one option for possible future improvements.

\section{Applications}

The models released as part of the MiDaS v3.1 family demonstrate high relative depth estimation accuracy with successful robustness and generalizability across environments. They are promising candidates for many applications---including architectures that combine relative and metric depth estimation~\cite{zoedepth, wofk2023videpth}, architectures for image synthesis~\cite{ho2020denoising, rombach2021highresolution, mildenhall2021nerf}, and architectures for text-to-RGBD generation~\cite{stan2023ldm3d, hollein2023text2room}.

\mypara{Metric depth estimation.} For practical applications requiring metric depth, MiDaS models on their own are insufficient as their depth outputs are accurate only up to scale and shift. Recent work has shown two approaches to resolving metric scale in depth outputs from MiDaS. Monocular visual-inertial depth estimation~\cite{wofk2023videpth} integrates generalizable depth models like MiDaS in conjuction with visual-inertial odometry to produce dense depth estimates with metric scale. The proposed pipeline performs global scale and shift alignment of non-metric depth maps against sparse metric depth, followed by learning-based dense alignment. The modular structure of the pipeline allows for different MiDaS models to be integrated, and the approach achieves improved metric depth accuracy when leveraging new MiDaS v3.1 models.

Whereas the above work relies on a combination of visual and inertial data, ZoeDepth~\cite{zoedepth} seeks to combine relative and metric depth estimation in a purely visual data-driven approach. The flagship model, ZoeD-M12-NK, incorporates a MiDaS v3.1 architecture with the BEiT-L encoder with a newly-proposed metric depth binning module that is appended to the decoder. Training combines relative depth training for the MiDaS architecture on the 5+12 dataset mix as described in~\cref{sec:training_setup}, followed by metric depth fine-tuning for the prediction heads in the bins module. Extensive results verify that ZoeDepth models benefit from relative depth training via MiDaS v3.1, enabling finetuning on two metric depth datasets at once (NYU Depth v2 and KITTI) as well as achieving unprecedented zero-shot generalization performance to a diverse set of unseen metric depth datasets.

\mypara{Depth-conditioned image diffusion.} MiDaS has been integrated into Stable Diffusion~\cite{rombach2021highresolution} in order to provide a shape-preserving stable diffusion model for image-to-image generation. Monocular relative depth outputs from MiDaS are used to condition the diffusion model to generate output samples that may vary in artistic style while maintaining semantic shapes seen in the input images. The depth-guided model released as part of Stable Diffusion v2.0 uses DPT-Hybrid from MiDaS v3.0 for monocular depth estimation. It is therefore very promising that MiDaS v3.1 models could be similarly integrated, with their improved depth estimation accuracy allowing for even better structure preservation in image-to-image diffusion.

\mypara{Joint image and depth diffusion.} Ongoing work in the text-to-image diffusion space has motivated the development of a Latent Diffusion Model for 3D (LDM3D)~\cite{stan2023ldm3d} that generates joint image and depth data from a given text prompt. To enable RGBD diffusion, LDM3D leverages a pretrained Stable Diffusion model that is fine-tuned on a dataset of tuples containing a caption, RGB image, and depth map. Training data is sampled from the LAION-400M dataset providing image-caption pairs. Depth maps corresponding to the images are obtained using DPT-Large from MiDaS v3.0. Supervised finetuning enables LDM3D to generate RGB and relative depth map pairs that allows for realistic and immersive 360-degree view generation from text prompts. Utilizing MiDaS v3.1 models to produce depth data for LDM3D finetuning could further improve the quality of LDM3D depth outputs and subsequent scene view generation.

\section{Conclusion}
We present a collection of robust depth estimation models in the new release MiDaS v3.1. Although we also explore convolutional backbones for the release, only transformer based backbones provide a sufficiently high depth estimation quality with the MiDaS architecture. The release v3.1 consists of depth models with the new transformer backbones BEiT, Swin, SwinV2, Next-ViT and LeViT, where we offer multiple different variants for BEiT and SwinV2. BEiT\textsubscript{512}-L with resolution 512x512 is on average 28\% more accurate than MiDaS v3.0 for non-square resolutions. The training of MiDaS has been extended from the original 10 datasets to 12, now including KITTI and NYU Depth V2 using the BTS split \cite{lee2019big}. For all of the released backbone types, we provide details on how they are integrated into the MiDaS architecture. We also consolidate this experience into a general guide to how MiDaS may be used with future backbones.

%\clearpage

%%%%%%%%% REFERENCES
{\small
\bibliographystyle{unsrt}
\bibliography{egbib}
}

\end{document}